# Thematic Analysis with Large Language Models: does it work with languages other than English? A targeted test in Italian

Stefano De Paoli – Abertay University (s.depaoli@abertay.ac.uk)

## Abstract

This paper proposes a test to perform Thematic Analysis (TA) with Large Language Model (LLM) on data which is in a different language than English. While there has been initial promising work on using pre-trained LLMs for TA on data in English, we lack any tests on whether these models can reasonably perform the same analysis with good quality in other language. In this paper a test will be proposed using an open access dataset of semi-structured interviews in Italian. The test shows that a pre-trained model can perform such a TA on the data, also using prompts in Italian. A comparative test shows the model capacity to produce themes which have a good resemblance with those produced independently by human researchers. The main implication of this study is that pre-trained LLMs may thus be suitable to support analysis in multilingual situations, so long as the language is supported by the model used.

## 1. Introduction

With recent prominence gained by Large Language Models such as GPT3.5. GPT4, LLama-2 or Mistral, the social sciences community has started to explore their application to qualitative data analysis. The use of LLMs for Thematic Analysis (TA), in particular, has shown good results, with solid propositions (see e.g. Author, 2023; Drapal et al., 2023) attempting to replicate most of the well know 6 phases of TA proposed by Braun and Clarke (2006) in their seminal work on this analysis method.

Thus far the existing publications have proposed performing TA with LLMs on data in English (e.g. semi-structured interviews in English). However, to be potentially viable as a method, the use of LLMs for TA should be assessed also with data in languages other than English. This may be particularly relevant for qualitative analysis, since for example qualitative interviews are often performed in national languages, studying local communities and with actors which do not necessarily speak English. Additionally, nuances of the language may be important for qualitative research and analysis. Moreover, this would better support multilingualism in social research, against strict dominance of English, or facilitate the work of researchers in the areas of language and multilingualism.

The goal of this paper is to present and discuss the results of conducting a TA supported by LLMs (specifically the commercially available models GPT3.5-Turbo and GPT4-Turbo), of data in Italian language. The choice of this language for the test is, first and foremost, practical, since the author speaks Italian and can therefore understand the nuances of the language and the data used (see later). Nonetheless, Italian constitutes a good case test for the purpose of the paper. It is a Romance/Latin language, thus sufficiently different from English. Moreover, it is a major European language, and it does appear a language supported by the LLMs used for this work (in other words the LLMs used were pre-trained also with material in Italian, albeit we can expect less than the equivalent English training data).

For performing this test, I will adopt a process similar to the one I proposed in previous papers, see Author (2023a, 2023b) and Author and Colleague (2024). This entails reproducing some of the key phases of TA proposed by Braun and Clarke (2006; 2021). In particular, in this paper I will perform the initial coding (phase 2), the definition of themes (phase 3), and a revision of the themes (phase 4). For the analysis I selected an Open Access dataset of semi-structured interview transcriptions by Gualandi et al. (2022) available from Zenodo, performed with arts and humanities scholars at the University of Bologna in Italy. The focus of the authors' research was to investigate the relations of humanities scholars with the term data, and to understand what they mean with 'data'. A related paper was also published by the authors (Gualandi et al., 2023). The Zenodo deposit also contains a file reporting the qualitative analysis (performed with the Open-Source software QualCoder) which was performed using Grounded Theory (GT). Although GT and TA are different methodologies, they share some common elements such as a phase of initial coding, and a second phase of identification of higher-level categories (like axial coding or identification of themes). The assumption is that we can use the coding analysis performed by the authors for some comparison at least. More details about the original analysis will be discussed in the methodology section of the paper. Ultimately, as previously written, the goal of performing an inductive TA with an LLM is not to reproduce exactly what a human analyst would do, but rather to deliver an analysis which is 'good enough' in capturing key patterns in a dataset. Two human analysts performing inductively a TA may, themselves, produce some different themes.

In the following pages the paper will present the following: first a brief review of the state-of-the-art of using LLM for qualitative analysis; second a thorough discussion of the methods and materials of the research work; third a presentation of the results covering initial codes and themes; fourth, an evaluation of the results to compare the LLM themes with the original categories produced by Gualandi et al. (2022). The paper will then present a discussion of the findings and a conclusion.

## 2. Literature review: LLM for qualitative analysis

Thematic Analysis is a widely diffused analysis procedure for qualitative textual data and other data formats. It is often adopted because of its flexibility and because it is generally theory agnostic and can be used within different forms of theorising (see e.g. Glasdamet al., 2023 or Glasdam and Stjernswärd, 2022 for two example using Goffman or Weber, and Aitken et al. 2019 using Bourdieu) and in a variety of social sciences disciplines and investigation areas. Its application includes research in e.g. medical and nursing research (see e.g. Vaismoradi et al., 2013 for an influential paper), social work (see Clark et al., 2023 for a recent contribution), sociology of work and labour (Nicholas and McDowall, 2012) or social media (Deighton-Smith and Bell, 2018) to mention a few examples. Braun and Clarke are generally considered the key authors of the approach and have written several influential papers which have popularised it, describing the method (2006; 2021), showing its application (e.g. 2018) and further contributing to other methodological aspects of the debate (2013; 2019), within the context of qualitative research. But other influential authors had written about the approach before them including Merton (1975) or Aronson (1994). For this paper the most important component of TA is that the approach proposed by Braun and Clarke in their seminal 2006 paper, sees the method perform 6 inter-related phases: 1) familiarisation with the data; 2) generation of initial codes; 3) grouping and sorting of codes into themes; 4) review of the themes; 5) summarisation and renaming of the themes; 6) write-up of the results.

With the recent and disruptive advent of Large Language Models, researchers have started to consider these models' capacity to perform qualitative forms of analysis. LLMs are AI systems working with Natural Language Processing. They are trained on a very large amount of textual material and can produce natural language conversations (e.g. in a chatbot like chatGPT), but can also perform other 'completion' tasks on textual content such as summarisation of text (see e.g. Wang et al., 2023), or identification of topics for example. In the field of textual analysis, LLMs have started to be used to perform certain forms of Qualitative Analysis. For example, this paper's author proposed a methodology which reproduces the 6 phases of TA by Braun and Clarke (see Author, 2023a, 2023b) with some success, performing the process entirely inductively. This approach uses the commercial OpenAI GPT models via the Application Program Interface (API) using python and passing the data (semi-structured interviews) programmatically to the LLM for the analysis. This same approach has further been tested by Mathis et al. (2024) with medical semi-structured interviews using another Open Source pre-trained LLM called Llama-2. Testing for this paper has shown a good degree of quality in the analysis by performing a comparison between a human made analysis and the LLM analysis, using semantic similarity and other techniques. Drapal et al. (2023) also proposed an emulation of phases 2-4 of TA with LLMs, using legal material in the form of facts descriptions from criminal court opinions as data, and also offering an evaluation comparing human and LLM codes. Author and Colleague (2024) also proposed a metric for measuring the saturation of initial coding (the phase 2 of the analysis), based on the ratio between the total number of initial codes and the number of unique (non-repeated codes). Zhang et al. (2023a) performed empirical research to assess how chatGPT can be used by researchers to refine Thematic Analysis. This uses the chatbot, rather than accessing programmatically the API as in the previously cited literature. Hamilton et al. (2023) also tested the use of chatGPT as support tool for performing qualitative thematic analysis, in this case also the chatbot is used rather than the API.

Other authors have focused only on the initial coding phase of the process, often employing a deductive approach, where codes are decided in advance and the model is asked to then use the codes to categorise the data. Works in this category for example include the papers by Gao e al. (2023) and Xiao et al. (2023). These deductive works often adopt inter-coder reliability between LLMs and human analysis to measure quality, which is possible because of the deductive nature of the process they adopt. Zhang et al. (2023b) also designed an early prototype tool supporting initial coding for a Thematic Analysis, even if they call these initial codes as themes. The works just cited have often a more software engineering-oriented approach than other works, with potentially less focus on discussing methodological issues that may arise when using LLM for qualitative analysis. Few other studies are performing other types of textual analysis. Sentiment analysis is, as one would expect, a quite popular approach (see for an applied example Zhang et al., 2023, or Carneros-Prado et al., 2023 for a study also encompassing emotion detection). Chew et al. (2023) have also proposed a process for deductive coding for content analysis with LLMs, whilst Huber and Carenini (2022), amongst others, offered reflections on discourse analysis with LLMs, in particular focusing on the structure of discourse.

As far as investigating the multilingual capacities (beyond the English dominance) of pre-trained LLMs in Thematic Analysis, the previously cited work by Drapal et al. (2023) used data (court documents) in Czech language. This appears so far, the only explicit case where data other than English was used for TA with LLMs. However, it does seem that the material was translated in English for performing the analysis, using automated translation. So even if the original data was in Czech, this was translated to English for the analysis. In any case, the authors' focus is not on testing the multilingual TA capacities of LLMs, and no reflections are therefore offered by them in this sense. In the field of evaluation of LLMs multilingual capacities (i.e. computing studies related to Natural Language Processing), there have been studies which have tried to assess the multilingual capacities of some LLMs (see for example Armengol-Estapé et al., 2021 for a study on Catalan on the GPT models, and Chang et al, 2023 for a survey which covers also existing literature on multilingual evaluations). These studies, however, are not focused on analysis in the social sciences, but more on an evaluation of the performance of the computational capacities of the models, when they deal with multilingual tasks. The following pages will propose therefore the study of performing TA in a language other than English with a test using data and prompt in Italian, as previously anticipated. This can be a first step toward addressing the gap of whether an LLM supported TA can be done with material in a variety of languages.

## 3. Method and Procedure

The process for performing the TA inductively with LLMs (the GPT3.5-Turbo and GPT4-Turbo models in this research), sees the use of python scripts to interact with the LLMs API in a programmatic fashion. In essence, python scripts are used to pass the data to an LLM, and prompts are used to instruct the model on the operations it should do on the data. The model will then return the response of the operation. Figure 1 summarises the process for the initial coding (phase 2 of TA) where the interviews are passed in the prompt within a python script which sends the interview and the instructions to the LLM API, which returns the set of initial codes formatted in JSON objects as response. The process is the same for the identification of themes, but for this it is the set of initial codes which is programmatically sent to the LLM, returning then a set of themes and the underlying codes.

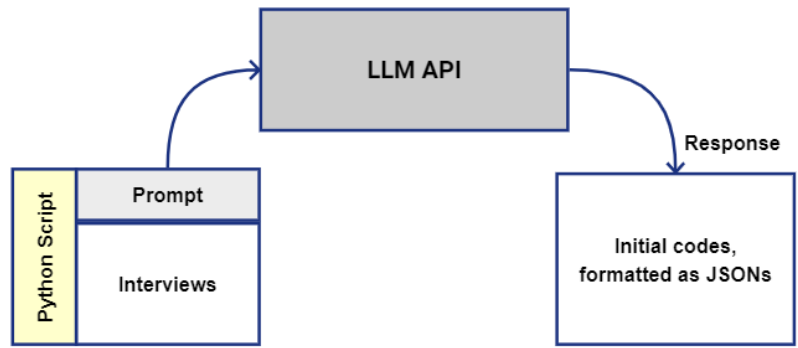

Figure 1 – Programmatic access of an LLM API for the Initial Coding

Key to these operations is the 'prompt', the set of instructions we provide to the LLM to perform actions on the data. A prompt in a chatbot like chatGPT would look like the one in Figure 2, where for example we ask the model to name us the 3 most important Italian poets of all time (the prompt), and the model gives us an output (the response). In the research accessing the API with python (like proposed here), the prompt is embedded in the python script, alongside other elements of code, which operate some data manipulation and parsing (like saving the initial codes in a .csv file). There are different strategies for prompting (see for an overview e.g. Wang et al., 2023), such zero-shot, few-shots, chain-of-thought. The one in the example relating to chatGPT (Figure 2) is zero-shot as it does not provide any example on which the model can produce a response, nor does it show the model how to think toward solving a problem. For TA so far, the author has used a zero-shot prompting strategy, which therefore relies entirely on the LLM capacity to produce a downstream output from a set of instructions.

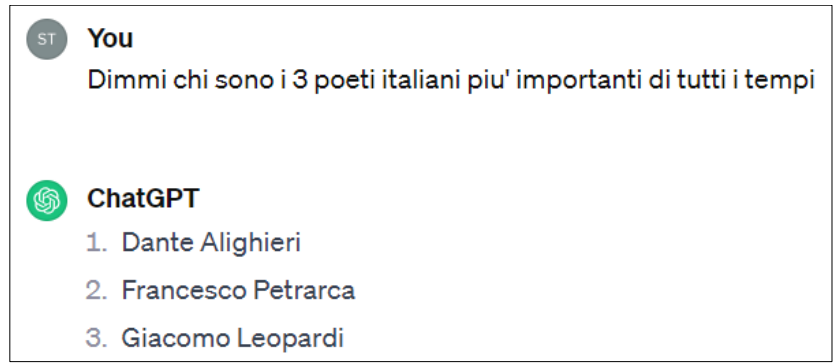

Figure 2 – Example of a simple prompt in chatGPT

For this research two prompts will be used: one for the initial coding (phase 2 of TA) and one for the identification of themes (phases 3-4 of TA). The prompts that are proposed here will also be in Italian, to test the capacity of the models used to understand instruction in the same language of the undelying dataset. One test will also be performed on initial coding (for one interview) using an equivalent prompt in English to assess whether the model derives similar initial codes when the instructions are on a different language than the dataset.

Figure 3 shows the prompt for the initial coding used within a python script. The response of the model to this prompt is captured by the get_completion function. This function has been described elsewhere (See e.g. Author, 2023) and readers can consult the previous papers. It is anyway a standard completion function for the GPT LLMs, widely document for example in the OpenAI website[1]. This prompt is the equivalent Italian translation of the prompts proposed in Author 2024. The prompt in Figure 3 (alongside the one shown in Figure 4) has some relevant features, compared with earlier prompts proposed for TA with LLMs. First, they are generally open-ended and do not fix in advance the number of initial codes to be generated. Second, the language used adheres to the jargon of Thematic Analysis, using key words such as 'initial codes' and 'themes' in the prompts, rather than more vague words such as topics or items. In the prompt each interview is passed in a variable ({testo}), and then to the model with the prompt (as summarised in Figure 1), which will return a list of initial codes (with names, descriptions and a representative quote), organised as a codebook. These are returned as JSON objects which are then easy to parse in a CSV file for the next step of the analysis, the generation of themes. For the generation of themes, the prompt in Figure 4 was used. The prompt receives the list of all initial codes and asks the model to generate a series of themes (at least 9, I fixed the themes number to match more or less the number of categories produced by Galandi et al., (2023) using the LLM generated codes. For performing phase 4 (the refining of themes), in previous work (see Author 2023a, 2023b) I proposed to modify the Temperature (T) parameter of the model. Temperature is a parameter that increases the 'creativity' of the LLM in responding to the prompt. Temperature at zero (0) normally delivers deterministic responses (always the same), whereas higher temperatures (say at 0.5 or 1) will increase the variability of the response, thus, in this case, leading to defining some potentially different themes, or offering different naming for similar themes.

---

[1] https://platform.openai.com/docs/quickstart?context=python

For phase 2 the model GPT3.5-Turbo was used, for phase 3 and 4 the model GPT4-Turbo was used. This second offers a larger context (i.e. it is possible to provide longer inputs and obtain longer outputs), but it would have been equally possible to use either model for each phase.

```
prompt = f"""

    Puoi assistermi nella generazione di una vasta gamma di categorie iniziali
    (genera tutte le categorie che ritieni indispensabili per catturare
    a  pieno il significato esplicito o latente, o gli eventi nel testo,
    concentrati sull'intervistato e non sull'intervistatore che fa le domande),
    L'obiettivo e' quello di raccogliere un ampio spettro di argomenti,
    azioni e idee presenti nel testo qui sotto, per aiutarmi nella conduzione
    di un'analisi tematica.

    Fornisci un nome per ciascuna categoria, con una descrizione densa di
    massimo 25 parole e una citazione dell'intervistato per ogni categoria di
    massimo 100 parole.

    Formatta la risposta come un file json mantenendo nomi, descrizioni e
    citazioni insieme nell'oggetto 'Categorie'.

    ```{testo}```
    """
response = get_completion(prompt)
print(response)
```

Figure 3 – Prompt in Italian for initial coding (equivalent to phase 2 in Braun and Clarke, 2006)

```
prompt = f"""

Leggi prima l'elenco delle categorie iniziali della mia
Analisi Tematica: {", ".join(codes_list)}.
Le categorie iniziali sono nel seguente formato:
[indice]: nome_codice. descrizione_codice. citazione

Determina tutti i possibili temi (almeno 9) ordinando, confrontando e
raggruppando le categorie iniziali.

Fornisci un numero adeguato di temi insieme a un nome, una descrizione densa
(120 parole) e l'elenco delle categorie (indice) per ciascun tema.
Assicurati che i temi catturino la ricchezza e la diversità dei codici iniziali.

"""
```

Figure 4 – Prompt in Italian for the identification of Themes (phase 3 of TA)

The data for this study is a set of qualitative interviews (n=19) with arts and humanities scholars conducted at the University of Bologna in Italy, with a research focus on understanding their relationship with data and what they mean and categorise as data. The dataset is available from Zenodo (Gualandi et al., 2023). The dataset also includes a file from the Qualitative Data Analysis software QualCoder, with the analysis categories. Although the work was performed using a grounded theory process, an inspection of the analysis shows a structure in high level categories, sub-categories and codes. I believe that these can still be used for some comparison with the TA generated by LLMs, to assess if LLMs can generate similar categories in the forms of themes. Table 1 shows the Categories taken from theQualCoder analysis file. They are in English in the original analysis (first two columns of the Table), and have been translated by me (second two columns) using chatGPT. Since the categories do not have a description, the sub-categories can be used to get a better idea of their focus.

| Categories (Highest Level) | Sub-Categories (or codes when sub-categories not available) | Categories (translated in Italian) | Sub-Categories (translated in Italian) |
|---|---|---|---|
| copyright and privacy | less-problematic materials; problematic materials | Diritto d'autore e Privacy | Materiali meno problematici; materiali problematici |
| general info on projects | challenges faced; how the project is funded | Informazioni generali sui progetti | Sfide affrontate; come il progetto è finanziato |
| is this data? and what is data? | data if…; opening up the definition of data; refusal or non applicability of term; restricting the definition of data | Questi sono dati? E cosa sono i dati? | Dati se...; ampliamento della definizione di dati; rifiuto o non applicabilità del termine; |

| | | | | restrizione della definizione di dati |
|---|---|---|---|---|
| materials | ancient manuscripts and early printed books; archival documents; born-digital artifacts; catalogues and other search tools; database created as part of a project; general-purpose databases used; subject-specific databases used; corpus; digital "infrastructures"; digital representations of cultural objects; documentation and standards; monuments, artworks and unique artifacts; personal data; publications, modern and contemporary; reading editions (not critical); critical editions; monographs and edited books; software; unpublished materials, modern and contemporary; websites | Materiali | Manoscritti antichi e libri a stampa; documenti archivistici; manufatti digitali nativi; cataloghi e altri strumenti di ricerca; database creati come parte di un progetto; database ad uso generale; database ad uso specifico del soggetto; corpus; "infrastrutture" digitali; rappresentazioni digitali di oggetti culturali; documentazione e standard; monumenti, opere d'arte e manufatti unici; dati personali; pubblicazioni, moderne e contemporanee; edizioni di lettura (non critiche); edizioni critiche; monografie e libri editi; software; materiali inediti, moderni e contemporanei; siti web |
| methodologies and workflows | keeping/discarding; philology and literature; language and linguistics; art history; computer science; archival studies; modes of collaboration; using existing hardware and software | Metodologie e flussi di lavoro | Conservare/eliminare; filologia e letteratura; lingua e linguistica; storia dell'arte; informatica; studi archivistici; modalità di collaborazione; utilizzo di hardware e software esistenti |
| research evaluation | ears around research ideas being stolen; not everything in research can be measured; research evaluation stifles collaboration and openess; research evaluation stifles innovation; the work of translators should be better recognized and valued; we need to find new ways of measuring research in the humanities | Valutazione della ricerca | Timori riguardo al furto di idee di ricerca; non tutto nella ricerca può essere misurato; la valutazione della ricerca ostacola la collaborazione e l'apertura; la valutazione della ricerca ostacola l'innovazione; il lavoro dei traduttori dovrebbe essere meglio riconosciuto e valorizzato; dobbiamo |

| | | | trovare nuovi modi per misurare la ricerca nelle discipline umanistiche |
|---|---|---|---|
| statements on open science, open data, open access | negative; neither; positive | Open Science, Open Data, Open Access - Dichiarazioni su Scienza Aperta, Dati Aperti, Accesso Aperto | Negativo; né negativo né positivo; positivo |
| statements on technology | importance of pen and paper; technology allowing advances in the discipline; technology changed the way of working | Dichiarazioni sulla Tecnologia | Importanza di carta e penna; tecnologia che permette progressi nella disciplina; la tecnologia ha cambiato il modo di lavorare |

Table 1 – Original categories from Gualandi et al. (2022) in the first two columns, and their translation

The categories by Gualandi et al. (2022) will be used to assess the results produced by the LLMs. For doing this I will propose an approach that uses another LLM (the model SBERT, see Thakur et al., 2022), specifically for semantic similarity. Semantic similarity is an approach to measure similarity between texts. It is a task of Natural Language Processing that measures the semantic 'distance' between two texts. In simpler terms the model will receive two 'texts' (e.g. an original category, and a theme from this research) and will produce a score for similarity or divergence, essential a score between 1 and –1 (cosine similarity). The closer the score is to 1 then the more similarity the sentences have. The closer to –1 then their meaning diverges toward being opposite. A score of 0 means that the texts are not similar at all. Using semantic similarity between the categories from the original analysis and the TA produced here, we could offer an assessment of how well the LLM performs with using semi-structured interviews and prompts in Italian. Results will be shown in a similarity matrix. This approach to assess the quality of the output was used (in a similar fashion) elsewhere in our work on TA with LLMs (see e.g. Mathis et al., 2024). The same approach will be used to compare the set of codes for one interview produced with the prompt in Italian and an equivalent prompt in English. This will be done to assess whether similar or equivalent codes are produced by the model when working with a prompt in a language different than the one of the underlying dataset.

A further evaluation will be proposed, based on a human comparative assessment of the themes and the original categories from Gualandi et al. (2022). For this I contacted the first author of the original research and asked to compare the LLM themes with their categories and to offer a qualitative score of similarity between 0-10, or in other words to score how well the themes from the model capture the meaning of the original categories. Results of this will also be shown in a matrix graph.

## 4. Results

In the following pages the results of the TA performed with the LLMs (GPT3.5-Turbo and GPT4-Turbo) will be presented. The results will follow the selected phases of TA, namely the initial coding, the identification of themes, and the refining of themes.

### 4.1 Initial coding

The initial coding of the interviews, performed with the prompt presented in Figure 2 and the GPT3.5-Turbo API, resulted in 185 codes (around 10 codes for each interview). Table 2 shows some examples of codes from the codebook which resulted from the analysis of the interviews, with their name, description and an illustrative quote. These are reported verbatim as produced by the model (with no changes made by the author of this paper). An English translation is also provided for each theme (in the light gray rows). Codes do appear of good quality, with simple yet illustrative names, a meaningful description and a quote that illustrates the codes. They are comparable in quality with codes seen in previous work on doing TA with LLM (see e.g. Author, 2023a), and therefore are usable for phase 3 of TA: the definition of a set of potential Themes.

| Code | Description | Quote |
|---|---|---|
| Edizioni digitali | Nuove frontiere nel settore delle edizioni digitali di fonti manoscritte | Il settore si sta aprendo con grande interesse anche con ricerche di carattere sperimentale alla possibilità di edizioni digitali. |
| Digital Editions | New frontiers in the field of digital editions of handwritten sources | The sector is opening up with great interest, including experimental research on the possibility of digital editions. |
| Metadati e standard di edizione | Standard di edizione critica e descrizione dei metadati delle fonti | L'edizione critica deve descrivere il metodo seguito e le scelte compiute, esplicitando il significato dei simboli e la struttura dell'edizione. |
| Metadata and Edition Standards | Critical edition standards and description of source metadata | The critical edition must describe the method followed and the choices made, explaining the meaning of symbols and the structure of the edition. |
| Dati della ricerca | I testi dell'edizione critica possono essere considerati dati. | L'edizione critica di un manoscritto puÃ² essere considerata un dato. |
| Research Data | The texts of the critical edition can be considered data. | The critical edition of a manuscript can be considered as data. |
| Privacy e sicurezza | Considerazioni sulla privacy dei dati raccolti e precauzioni etiche prese. | Ci sono aspetti di privacy sicuramente nei dati: raccogliendo dati sulle persone c'è questo aspetto, c'è un task dedicato a questo nel progetto. |
| Privacy and Security | Considerations on the privacy of collected data and ethical precautions taken. | There are certainly privacy aspects in the data: when collecting data about individuals, this aspect is present, and there is a dedicated task to address this in the project. |
| Condivisione dei materiali | Chi ha accesso ai materiali e se vengono condivisi con colleghi o il pubblico. | La condivisione avviene ma ad esempio nella correzione delle bozze, quindi tipicamente noi ci leggiamo fra noi le cose. |
| Sharing of Materials | Who has access to the materials and whether they are shared with colleagues or the public | Sharing takes place, but for example, in the proofreading phase, so typically we read things among ourselves. |

Table 2- Examples of codes generated by the LLM with the prompt in Figure 2

## 4.2 Themes

Phase 3 of TA encompasses the sorting and organising of initial codes into higher level Themes. These are the patterns that emerge from the data and in essence the core analytical component of the TA method. Using the prompt in Figure 3 and passing the list of the initial codes previously generated as input in the prompt, the model (GPT4-Turbo in this phase) can define these themes. The prompt request to generate a set of themes with a request to have at least 9. The reason for this is that the original research produced 8 high-level categories. With 9 themes from the model, we can better see if some of these themes do appear similar to the original ones, having one extra theme to play with. Table 3 presents the 9 Themes produced (reported verbatim and in the same order given by the model, in appendix an English translation produced with chatGPT is provided). We can see that the model offers themes with a dense description. Examples of codes included in each theme are reported in the last column. In the appendix the translation of two themes is reported to allow better judgement of the quality of this output.

| Theme | Description | Examples of codes included |
|---|---|---|
| Tema 1: Metodologie e Standard di Ricerca | Questo tema esplora le diverse metodologie e standard adottati nei progetti di ricerca umanistica. Include l'approccio multisciplinare, l'uso di tecnologie come la modellazione 3D, le pratiche di edizione critica, e le metodologie di analisi testuale e visiva. Si considerano anche le pratiche di conservazione digitale e fisica, l'importanza delle buone pratiche e la sistematizzazione delle metodologie, nonché la definizione di standard riconosciuti a livello internazionale. | Metodologie riconosciute nella disciplina; Metadati e standard di edizione |
| Tema 2: Conservazione e | Questo tema si concentra sulla conservazione e l'accessibilità dei materiali di ricerca, sia in formato digitale che cartaceo. Tratta della gestione dei materiali nei musei e archivi, delle politiche di accesso | Formato e conservazione dei materiali; |
| Accesso ai Materiali | aperto, delle restrizioni imposte da musei o istituzioni, e della tendenza a conservare tutto il materiale di ricerca per un uso futuro. Si esplora anche la condivisione dei materiali tra ricercatori e la pubblicazione finale per il pubblico. | Documentazione dei materiali |
| Tema 3: Proprietà Intellettuale e Privacy | Questo tema affronta le questioni di proprietà intellettuale, copyright e privacy nei progetti di ricerca umanistica. Si discute della donazione di modelli 3D ai musei, delle preoccupazioni per il copyright delle foto dei manoscritti, della gestione di informazioni confidenziali e della privacy dei dati raccolti. Si considera anche l'impatto di questi aspetti sulla condivisione dei risultati della ricerca e sulla pubblicazione di materiali. | Privacy e sicurezza; Informazioni confidenziali e proprietà intellettuale |
| Tema 4: Dati e Definizioni nella Ricerca Umanistica | Questo tema esplora la definizione e il ruolo dei dati nella ricerca umanistica. Si indaga come i dati vengono raccolti, interpretati e utilizzati per costruire nuove conoscenze. Include la considerazione di testi, manoscritti, iscrizioni e altri materiali come dati, nonché la trasformazione di testi in dati attraverso metodologie innovative. Si discute anche della definizione personale di dati da parte dei ricercatori e della loro importanza per la creazione di nuove storie della letteratura e della cultura. | Definizione di 'dati'; Dati della ricerca |
| Tema 5: Open Science e Open Data | Questo tema si concentra sull'importanza dell'open science e dell'open data nel campo umanistico. Si discute della necessità di una maggiore attenzione all'open science, della possibilità di creare nuove storie della letteratura attraverso la liberalizzazione delle fonti grezze, e della condivisione gratuita dei dati di ricerca. Si esplora anche la considerazione di tutto ciò che è misurabile o digitalizzabile come dato e la potenzialità di questi dati per la comunità scientifica e il pubblico più ampio. | Open science e dati umanistici; Open data |
| Tema 6: Progetti di Ricerca e Collaborazione | Questo tema riguarda la varietà e la portata dei progetti di ricerca umanistica, evidenziando la collaborazione tra ricercatori di diverse discipline e istituzioni. Si esplorano progetti che vanno dall'edizione critica di testi antichi alla ricerca iconografico-stilistica, dalla traduzione letteraria alla linguistica acquisizionale. Si considera anche l'impatto della collaborazione e della condivisione dei materiali e dei risultati sul progresso della ricerca. | Progetto di edizione critica delle lettere manoscritte; Progetti di ricerca |
| Tema 7: Materiali di Ricerca e Fonti | Questo tema si concentra sui diversi tipi di materiali e fonti utilizzati nella ricerca umanistica, come manoscritti, iscrizioni, fonti primarie e secondarie, e dati digitali. Si esplora la raccolta, l'analisi e l'uso di questi materiali per la produzione di nuove conoscenze, nonché le sfide legate alla loro conservazione e accessibilità. Si considera anche l'uso di banche dati, archivi online e piattaforme digitali per la ricerca bibliografica e la gestione dei materiali. | Descrizione dei materiali; Materiali di ricerca |
| Tema 8: Metadati, Edizioni e Pubblicazioni | Questo tema esplora l'importanza dei metadati, delle edizioni critiche e delle pubblicazioni nella ricerca umanistica. Si discute della creazione di edizioni scientifiche e divulgative, dell'uso di standard di edizione critica, e della descrizione dei metadati delle fonti. Si considera anche la pubblicazione di risultati di ricerca come dati aperti e la potenzialità delle edizioni digitali per la disseminazione della conoscenza. | Metadati e standard di edizione; Progetto di traduzione di apocrifi cristiani |

| Tema 9: Tecnologie e Innovazione nella Ricerca Umanistica | Questo tema si concentra sull'uso delle tecnologie e dell'innovazione nei progetti di ricerca umanistica. Include l'adozione di software per la gestione dei dati, l'uso di piattaforme digitali per la Citizen Curation, e lo sviluppo di app e strumenti online per l'apprendimento delle lingue. Si esplora anche l'uso di tecnologie come la modellazione 3D per lo studio di iscrizioni antiche e la creazione di banche dati digitali per l'estrazione di metadati dai documenti. | Documentazione e standard; Banca dati digitale |
|---|---|---|

Table 3 – Themes generated by the model using the initial codes

## 4.3 Refining themes

In previous work (see e.g. Author, 2023a) I proposed to change the Temperature of the model to perform phase 4 of TA, to refine and identify themes that might have not been identified in phase 3. This was discussed in the paper's methodology before. For this, the prompt proposed in Figure 3 was run 3 times with T at 0.25, 0.5 and 0.75, to identify themes overlooked in phase 3 (Table 4). As we can see there is a general consistency across Table 3 and Table 4 (which presents the 3 iterations) with most themes overlapping. Only 1 theme (last row Table 4) for each set does not appear in Table 3. These, however, taken individually do not appear elsewhere in Table 4, therefore they may not appear relevant themes. In Table 3 one of the Themes does not appear in Table 4 anywhere, specifically the theme "Metadati, Edizioni e Pubblicazioni" (Metadata, Editions and Publications). This also may not be a relevant theme.

| **T - 0.25** | **T - 0.5** | **T - 0.75** |
|---|---|---|
| Metodologie e Pratiche di Ricerca | Metodologie e Standard di Ricerca | Metodi di Ricerca e Multidisciplinarietà |
| Accesso e Condivisione dei Materiali | Conservazione e Formato dei Materiali | Conservazione e Accessibilità dei Materiali |
| Proprietà Intellettuale e Copyright | Accesso e Condivisione dei Materiali | Proprietà Intellettuale e Copyright |
| Conservazione e Formato dei Materiali | Proprietà Intellettuale e Copyright | Metodologie e Standard nella Ricerca |
| Dati e Definizione di Dati nella Ricerca | Dati della Ricerca | Open Science e Accesso ai Dati |
| Tecnologie Digitali e Innovazione | Progetti di Ricerca Specifici | Tecnologie Digitali e Innovazione |
| Collaborazione e Scambio di Informazioni | Materiali di Ricerca e Loro Utilizzo | Collaborazione e Condivisione nella Ricerca |
| Privacy e Sicurezza dei Dati | Innovazioni Tecnologiche e Digitalizzazione | Privacy, Sicurezza e Etica nella Ricerca |
| Pubblicazione e Disseminazione dei Risultati | Pratiche di Ricerca e Insegnamento | Conservazione Storica e Tradizione |

Table 4 – Themes generated with different T (0.25, 0.5, 0.75)

## 5. Evaluation of the results

Table 5 puts side by side the high-level categories from the original research by Gualandi et al. (2022) and the themes produced by the LLM. These have been paired manually for initial similarity by the author, with the last row only reporting categories/codes that do not seem to be similar, where the category 'research evaluation' for example never emerged as Theme in either phase 3 or 4 using the LLM. A comparison will be performed on the last two columns for semantic similarity. In a couple of cases for the original categories translation some small adjustments were made to make latent meaning more explicit, or few details were included taken from the underlying sub-categories. For example, the original 'general info on projects' was translated as 'Informazioni generali sui progetti di ricerca' (general information on research projects), with the term research added as the categories did refer to the research projects in which researchers work or have worked, and so the latent meaning was made explicit. The original category 'is this data? and what is data?' was perhaps evocative for the original analysis, but it hides the consideration that this code was trying to capture aspects of the analysis related to defining what data is in the arts and humanities. To offer some more clarity for similarity, the sub-categories of the original analysis were included for augmenting the clarity of this category when performing semantic similarity. Likewise, for the original category 'statements on technology', the underlying sub-categories are included, in order to clarify the meaning of this a bit better.

| Original (Gualandi et al., 2022) | LLM Themes reordered for similarity | Original translated in Italia and adjusted |
|---|---|---|
| copyright and privacy | Tema 3: Proprietà Intellettuale e Privacy | Diritto d'autore e Privacy |
| general info on projects | Tema 6: Progetti di Ricerca e Collaborazione | Informazioni generali sui progetti di ricerca |
| is this data? and what is data? | Tema 4: Dati e Definizioni nella Ricerca Umanistica | Questi sono dati? E cosa sono i dati? - ampliamento della definizione di dati; rifiuto o non applicabilità del termine dati; restrizione della definizione di dati |
| materials | Tema 7: Materiali di Ricerca e Fonti | Materiali |
| methodologies and workflows | Tema 1: Metodologie e Standard di Ricerca | Metodologie e flussi di lavoro |
| statements on open science, open data, open access | Tema 5: Open Science e Open Data | Dichiarazioni su Open Science, Open Data, Open Access |
| statements on technology | Tema 9: Tecnologie e Innovazione nella Ricerca Umanistica | Dichiarazioni sulla Tecnologia - tecnologia che permette progressi nella ricerca umanistica e arti; la |

|  |  | tecnologia ha cambiato il modo di lavorare |
|---|---|---|
| research evaluation | Tema 8: Metadati, Edizioni e Pubblicazioni<br><br>Tema 2: Conservazione e Accesso ai Materiali | [not used for similarity test] |

Table 5 – Original categories and LLM themes

Figure 4 shows the results of the semantic similarity. The similarity matrix shows on the vertical side the original categories and on the horizontal side the Themes from the LLM. The diagonal shows the key scores we need to consider, and they report the result of the cosine similarity when the original categories and the LLM Themes are compared. For 6 out of the 7 comparisons the diagonal shows a cosine similarity of 0.6 and above, which is encouraging and overall, it shows that the model can identify with some degree of validity similar categories as the human analyst. Only for the category/theme pair related to technology, does there seem to be limited similarity (at 0.43).

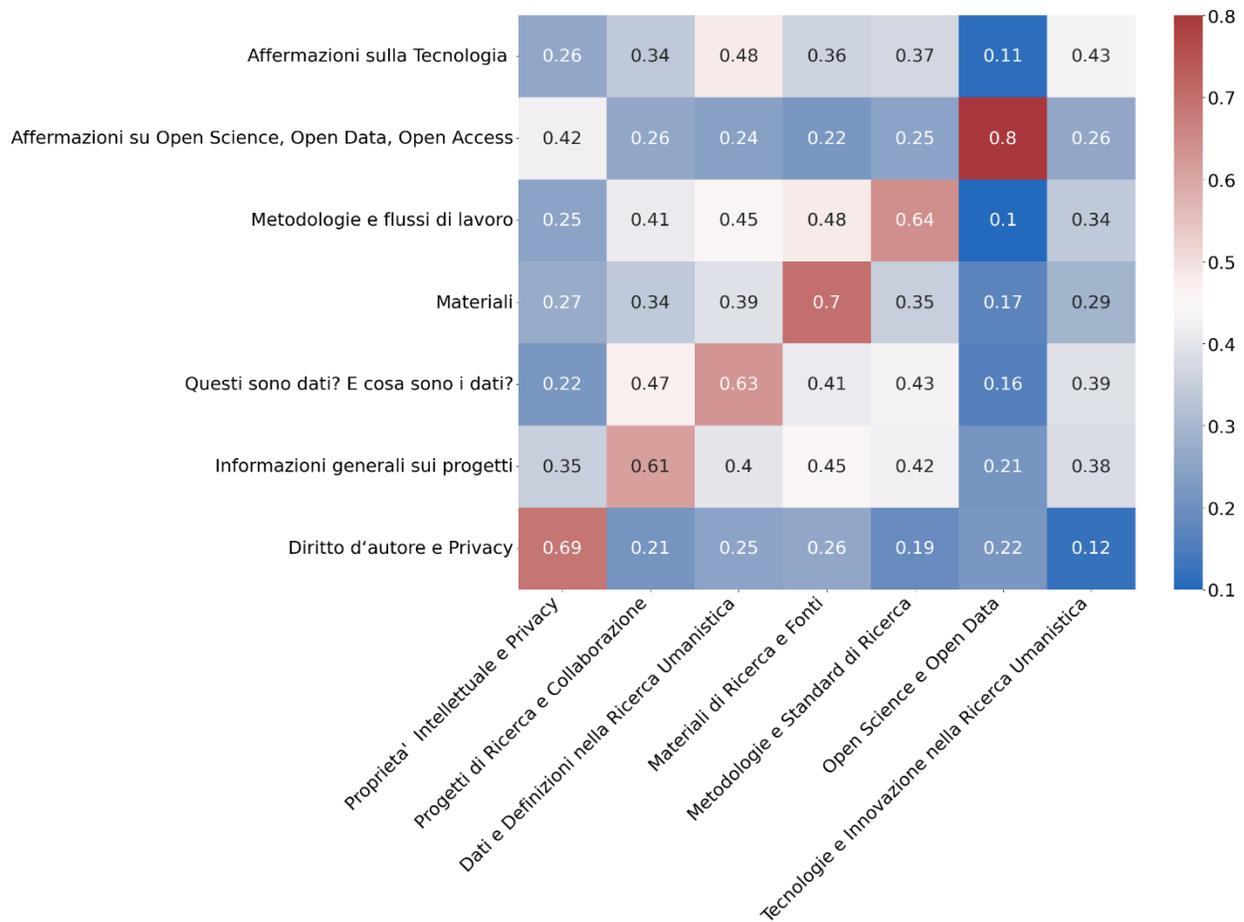

Figure 4 – Similarity scores between original categories and themes (only based on their names)

As the original categories often had some latent meaning, a further semantic similarity test was done by modifying the original category from "affermazioni sulla tecnologia", to "affermazioni sulla tecnologia digitale", just adding therefore a clarification that what is talked about is digital technology, and not generically technology. Moreover, the theme from the LLM used for the additional comparison is taken from one the phase 4 lists, "Tecnologie Digitali e Innovazione", appearing in both the Themes with T at 0.25 and 0.75. We can see in Figure 5 that the semantic similarity is much more significant, this we could assume that the LLMs can identify something close to the original category in what is the equivalent of the phase 4 of TA.

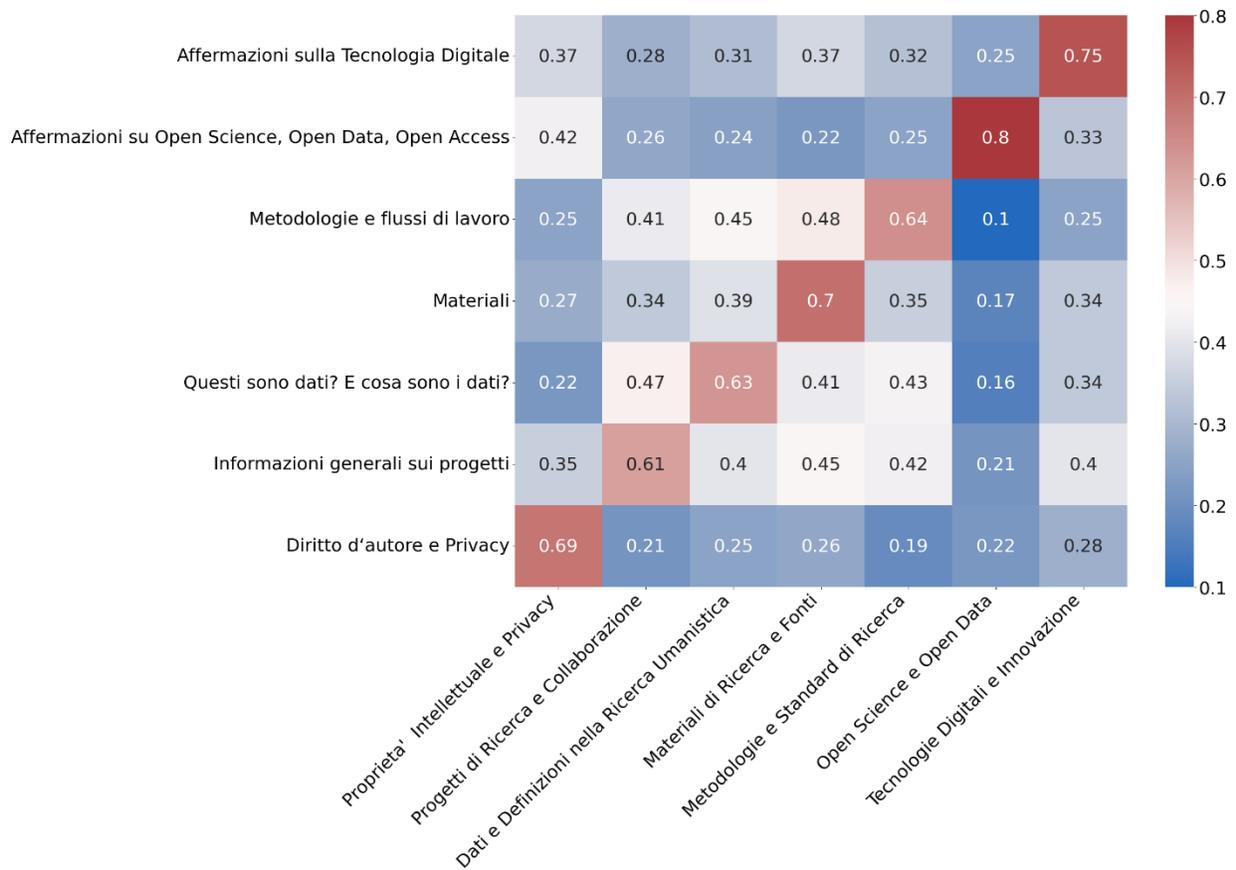

Figure 5 - Similarity scores between original categories and themes (only based on their names), changing the last pair

One last automated similarity test was performed on two sets of codes. The codes produced by the model for 1 interview (in this case the interview nr 14 of the original dataset) are compared to the codes produced using the prompt in the original language of the data, and an equivalent prompt in English which also asks to produce the codes and description "in the original language of the interview data". This is to test whether the language of the prompt might have any very significant effect on the codes produced. The list of codes in the second column of Table 6 is the one performed with the prompt in Italian. The codes in the third column were produced with the prompt in English. The codes have been re-ordered for perceived similarity by me. Clearly, the two prompts produce similar codes, where the codes 1-12 are often identical. Some of the codes from the prompt in Italian do appear slightly more detailed from example the first pair of codes where "Progetto di edizione critica delle lettere manoscritte" contained more detils about the research, than simply the code "Progetto di ricerca".

|    | **Codes with Prompt in Italian** | **Codes with Equivalent Prompt in English (forcing names in original language)** |
| --- | --- | --- |
| 1  | Progetto di edizione critica delle lettere manoscritte | Progetto di ricerca |
| 2  | Materiali di ricerca | Materiali di ricerca |
| 3  | Edizione critica | Edizione critica |
| 4  | Banca dati digitale | Banca dati digitale |
| 5  | Conservazione dei materiali | Conservazione dei materiali |
| 6  | Accesso ai materiali | Accesso ai materiali |
| 7  | Condivisione dei materiali | Condivisione dei materiali |
| 8  | Informazioni confidenziali e privacy | Informazioni confidenziali |
| 9  | Dati della ricerca | Dati della ricerca |
| 10 | Conservazione dei materiali di lavoro | Conservazione dei materiali di lavoro |
| 11 | Open data | Open data |
| 12 | Definizione di dati | Definizione di dati |
| 13 | Proprietà intellettuale e copyright | Pubblico dei materiali |
| 14 | Metadati e standard di edizione | Formato dei materiali |
| 15 | Edizioni digitali |  |

Table 6 – Set of codes from one interview produced with the prompt in Italian and an equivalent in English

Figure 6 shows the results of the semantic similarity on this set of codes, with the codes created with the prompt in Italian on the vertical side of the matrix. The last two codes on the upper right corner show low similarity scores as anticipated since in rows 13 and 14 of Table 5 we have paired codes, that did not have similar equivalent. The first pair of codes in the bottom left corner, also show low similarity scores. But in general, for the large majority of codes, the prompts in Italian (the original language of the data) and in in English (forcing to give codes in the language of the data) perform in a similar fashion. Possibly they may be used interchangeably at the researcher's discretion.

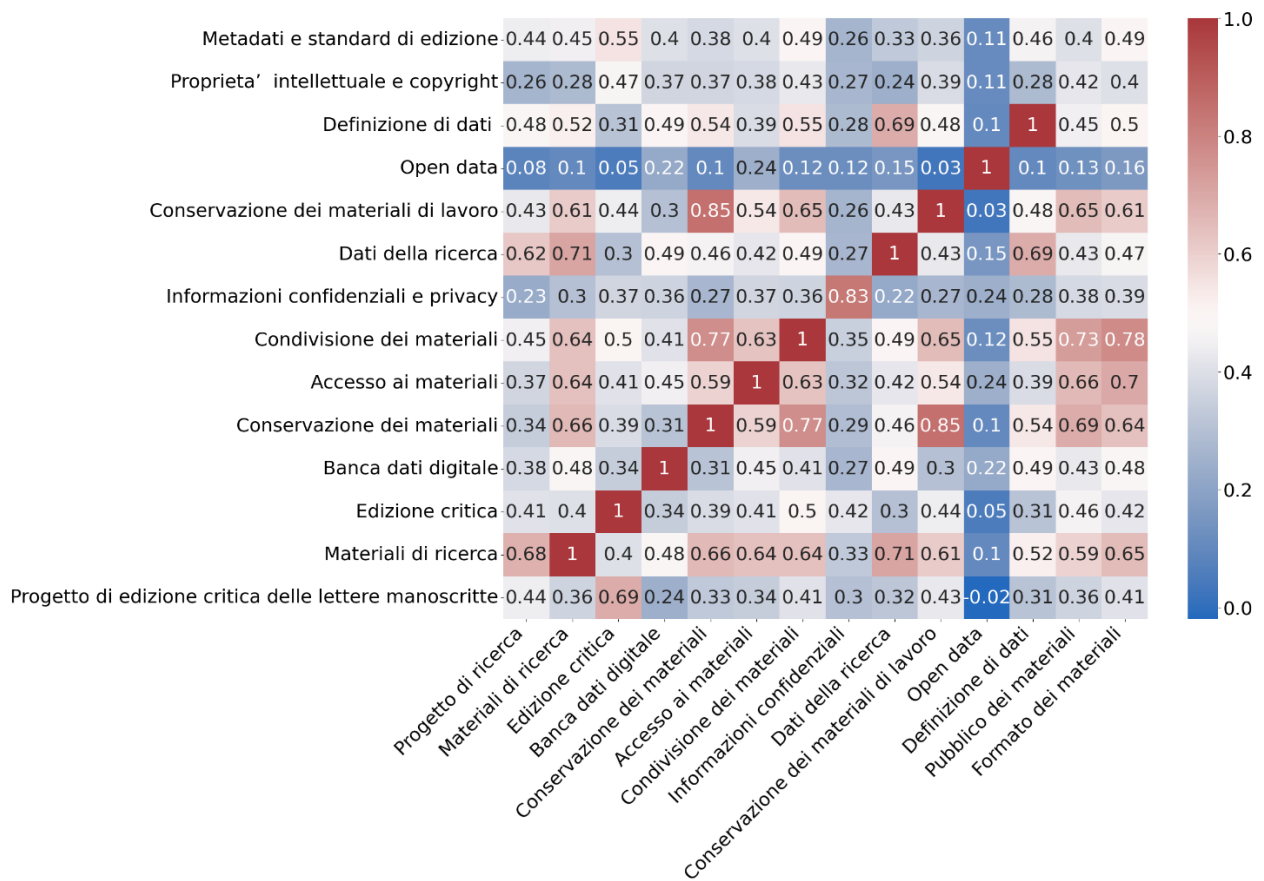

Figure 6 – Semantic similarity between codes produced with prompt in Italian and equivalent in English

Figure 7 shows the scores obtained from the lead author of the original research. As discussed in the methods we asked the first author of the original research to put a score between 0 to 10 (which is equivalent to scoring between 0 and 1) to evaluate how well the LLM themes do capture the meaning of the original categories. For this we used the same pair category-theme seen in Figure 5. As we can see the scores provided are quite high, with in two cases an assessment that the themes capture in full the meaning of the human produced categories (score 10), but even in the case of the other remaining 4 categories the scores are 8s and 9s. This further consolidates the previous observations and confirms the capacity of the pre-trained LLMs used in this research, to work with material in Italian and to identify themes which are of similar quality as that of human researchers.

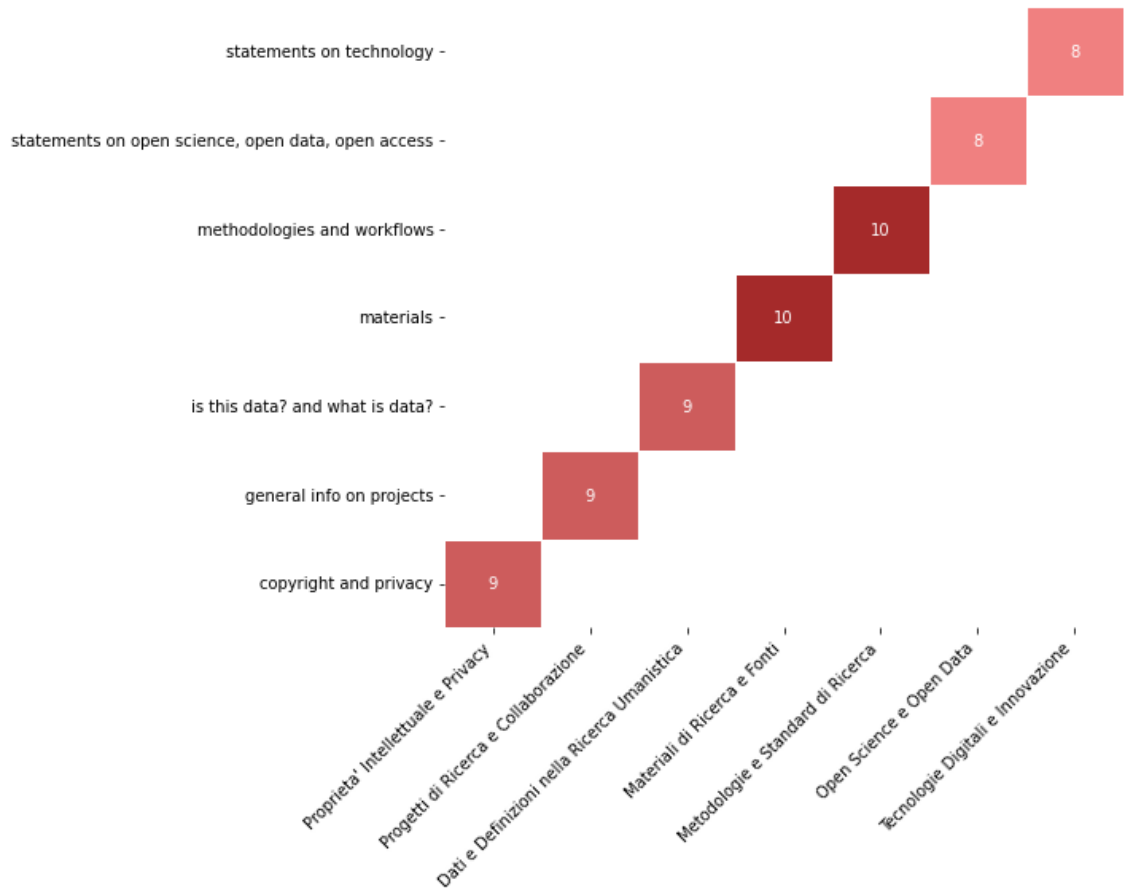

Figure 7 – Scores provided by the author of the original research categories/LLM themes

## 6. Discussion and Conclusion

The problem of this paper was to test whether an inductive Thematic Analysis performed with a pre-trained Large Language Model can be performed on semi-structured interviews in a language other than English. For this purpose, a test was conducted using a set of qualitative semi-structured interviews in Italian. Moreover, for the test, the methodology for TA with LLMs proposed previously by the author (see e.g. Author, 2023a) was adopted. The question is then, does the LLM perform a TA which is 'good enough'? That displays sufficient quality? Even when LLMs work with material other than English.

Overall, we have seen how the pre-trained models (GPT3.5-Turbo and GPT4-Turbo), can identify a set of codes from each of the interviews in Italian. These codes do appear of sufficient quality. The model has identified 185 codes from the 19 interviews. These were then used within a prompt to instruct the model to identify the themes. In the research we asked to identify 9 themes. These were then compared with the 8 high-level categories (created with Grounded Theory) produced by Gualandi et al. (2022, 2023) in their original analysis of the interviews, for semantic similarity.

The LLMs can produce themes which are generally comparable in the large majority of cases. Only one of the original categories by Gualandi et al. (2022) was never inductively produced by the model (the category 'research evaluation'), but the other 7 were generally seen by the LLM, showing semantic similarity (derived by scoring pairs categories/themes with sbert) with a score always >0.6. For this test, moreover, the prompts for the LLM were used in the original language of the dataset. A test for the themes was done with an equivalent prompt in English, showing that the original language and the English prompt can generally produce similar themes. A further quality assessment was performed by asking one of the authors of the original research to assign a score about how well the LLM themes capture the significance and meaning of the original categories. In this case the scoring also appeared rather high, thus further confirming the LLMs capacity to perform an inductive qualitative analysis.

This research consolidates the existing initial body of work on using LLMs for TA, which was discussed in the literature review. There is no direct innovation in the procedure for the analysis described elsewhere (see e.g. Author 2023; Mathis et al., 2024), however there are important practical implications coming from this research. First, it shows (at least in one example case – Italian) that pre-trained LLMs have multilingual capacity to perform the analysis. This would allow the methodology to be adopted for example in international projects (like European projects) where often qualitative data (such as semi-structured interviews) is collected in diverse national languages, supporting a common analysis procedure without the need for translating interview in English for example, or allowing a common analysis even if individual researchers from different countries perform their own analysis in parallel. Moreover, as qualitative research seeks to explore the deep understanding of social issues from the perspective of the actors, the multilingual capacities of LLMs could allow the methodology to be adopted in multiple linguistic contexts. This of course, so long that the language is supported by the pre-trained model, in a similar fashion as in the test with Italian proposed in this work.

In this research we used pre-trained commercial models which are easy to access and (at least for GPT3.5.-Turbo) are relatively cheap to use. But the scientific community is also fine-tuning Open-Source models in specific languages. For example, for Italian a large model based on the Llama-2 70b was recently released, called "Llamantino" (Basile et al., 2023). This is instruction tuned for supporting Italian NLP researchers. However, the general overhead in terms of computing resources or technical skills needed to run (at least at present) an Open-Source model such as this are quite high. An Open Source model the size of Llamantino would require dedicated hardware resources and technical mastery not always available to social scientists. Whilst in the future these fine-tuned models in a specific language should be tested in their capacity to perform TA (using the methods discussed earlier), at present the commercially available models do appear easier to use, and do not require any significant investment in technology, nor they require very advanced computing skills (beyond some basic python programming). Further incremental research should possibly test the TA with LLMs as described here with other languages to further validate the findings. The choice to use Italian was practical (because the author knows the language), and it was assessed that Italian is still a major European language. Tests may be performed with other major languages (e.g. German) but possibly also with languages of smaller countries (e.g. Croatian or Slovak for example).

In conclusion a TA on semi-structured interviews in Italian can be performed with a commercial, pre-trained LLM. The analysis results do appear of sufficient 'good quality', generally with some comparability to the original research performed by human researchers. Moreover, the outputs produced by the model do appear similar to those produced in previous research (e.g. Author, 2023) on data in English, also using prompts in English. Performing a TA with LLM in Italian thus does seem viable, and this contribution potentially opens reflections on a broader multilingual application of the method.

**Acknowledgement.** The author would like to thank Bianca Gualandi from the University of Bologna for the support received in the evaluation of the results.

# Appendix

**English translation of the Themes generated by the LLM**

| Theme | Description | Example of code |
| --- | --- | --- |
| Theme 1: Methodologies and Research Standards | This theme explores the various methodologies and standards adopted in humanities research projects. It includes a multidisciplinary approach, the use of technologies such as 3D modeling, critical editing practices, and textual and visual analysis methodologies. It also considers digital and physical conservation practices, the importance of best practices and the systematization of methodologies, as well as the definition of internationally recognized standards. | Recognized methodologies in the discipline; Metadata and editing standards |
| Theme 2: Conservation and Access to Materials | This theme focuses on the conservation and accessibility of research materials, both in digital and paper formats. It deals with material management in museums and archives, open access policies, restrictions imposed by museums or institutions, and the tendency to preserve all research material for future use. It also explores material sharing among researchers and final publication for the public. | Material format and conservation; Material documentation |
| Theme 3: Intellectual Property and Privacy | This theme addresses intellectual property, copyright, and privacy issues in humanities research projects. It discusses the donation of 3D models to museums, concerns regarding copyright of manuscript photos, management of confidential information, and data privacy. It also considers the impact of these aspects on research results sharing and material publication. | Privacy and security; Confidential information and intellectual property |
| Theme 4: Data and Definitions in Humanities Research | This theme explores the definition and role of data in humanities research. It investigates how data is collected, interpreted, and used to build new knowledge. It includes the consideration of texts, manuscripts, inscriptions, and other materials as data, as well as the transformation of texts into data through innovative methodologies. It also discusses researchers' personal definition of data and its importance in creating new literary and cultural narratives. | Definition of 'data'; Research data |
| Theme 5: Open Science and Open Data | This theme focuses on the importance of open science and open data in the humanities field. It discusses the need for greater attention to open science, the possibility of creating new literary narratives through the liberalization of raw sources, and the free sharing of research data. It also explores the consideration of anything measurable or digitizable as data and the potential of this data for the scientific community and the wider public. | Open science and humanities data; Open data |
| Theme 6: Research Projects and Collaboration | This theme concerns the variety and scope of humanities research projects, highlighting collaboration among researchers from different disciplines and institutions. It explores projects ranging from critical editing of ancient texts to iconographic-stylistic research, from literary translation to acquisitional linguistics. It also considers the impact of collaboration and material and results sharing on research progress. | Critical edition project of handwritten letters; Research projects |
| Theme 7: Research Materials and Sources | This theme focuses on the different types of materials and sources used in humanities research, such as manuscripts, inscriptions, primary and secondary sources, and digital data. It explores the collection, analysis, and use of these materials for the production of new knowledge, as well as the challenges related to their conservation and accessibility. It also considers the use of databases, online archives, and digital platforms for bibliographic research and material management. | Material description; Research materials |
| Theme 8: Metadata, Editions, and Publications | This theme explores the importance of metadata, critical editions, and publications in humanities research. It discusses the creation of scientific and popular editions, the use of critical edition standards, and the description of metadata from sources. It also considers the publication of research results as open data and the potential of digital editions for knowledge dissemination. | Metadata and edition standards; Project for translation of Christian apocrypha |
| Theme 9: Technologies and Innovation in Humanities Research | This theme focuses on the use of technologies and innovation in humanities research projects. It includes the adoption of software for data management, the use of digital platforms for Citizen Curation, and the development of apps and online tools for language learning. It also explores the use of technologies such as 3D modeling for the study of ancient inscriptions and the creation of digital databases for metadata extraction from documents. | Documentation and standards; Digital database |